\pdfoutput=1
\let\realTexYear\year         
\documentclass{ieeeaccess}    
\let\ieeeYear\year            
\let\year\realTexYear         

\usepackage{xcolor}
\definecolor{accessblue}{rgb}{0.0,0.28,0.73}

\usepackage{cite}
\usepackage{amsmath,amssymb,amsfonts}
\usepackage{algorithmic}
\usepackage{graphicx}
\usepackage{textcomp}
\usepackage{url}
\usepackage{tikz}
\usetikzlibrary{shapes, arrows.meta, positioning, calc, shadows, decorations.pathreplacing, patterns, backgrounds}
\usepackage{tabularx}

\pgfdeclarelayer{background}
\pgfdeclarelayer{foreground}
\pgfsetlayers{background,main,foreground}

\def\BibTeX{{\rm B\kern-.05em{\sc i\kern-.025em b}\kern-.08em
    T\kern-.1667em\lower.7ex\hbox{E}\kern-.125emX}}
\usepackage{xurl}

\begin{document}
\history{Date of publication Jan 7, 2026, date of current version Jan. 7, 2026.}
\doi{10.1109/ACCESS.2025.0000000}

\title{Generative AI and Digital Ecosystem Resilience: A Proactive Lifecycle-Based Survey}
\author{\uppercase{Jonghyun Chung}\authorrefmark{1}, 
\uppercase{Rishabh Chaddha}\authorrefmark{1},
\uppercase{Sanket Badhe}\authorrefmark{1},
\uppercase{Debanshu Das}\authorrefmark{1},
\uppercase{Nathan Huang}\authorrefmark{1},
\uppercase{Amanpreet Kaur}\authorrefmark{1}}

\address[1]{Google LLC}

\markboth
{Author \headeretal: Generative AI and Digital Ecosystem Resilience: A Proactive Lifecycle-Based Survey}
{Author \headeretal: Generative AI and Digital Ecosystem Resilience: A Proactive Lifecycle-Based Survey}

\corresp{Corresponding author: Jonghyun Chung (e-mail: johnchung@google.com).}

\begin{abstract}
The proliferation of adversarial synthetic content, accelerated by Generative AI (GenAI) is rendering traditional reactive detection methods ineffective. This survey synthesizes emerging research to demonstrate a paradigm shift toward the proactive detection of emerging inauthentic narratives. In this survey, we adopt a unified, lifecycle-based taxonomy to combine socio-technical lifecycle models of adversarial campaigns with advanced computational methodologies for emerging inauthentic narrative detection. By structuring the analysis around the C5 Interaction Model (Context, Causes, Content, Cycle of Amplification, Consequences), we integrate different research streams from machine learning and social science. To differentiate spread patterns of synthetic amplification from authentic baseline traffic, this paper surveys state-of-the-art techniques for modeling the creation, seeding, and propagation of fresh narratives, including the analysis of Coordinated Inauthentic Behavior (CIB), epidemiological modeling, and Hawkes process. This survey also provides a systematic review of proactive detection methods for adversarial threats at different stages in the C5 interaction model—specifically, anomaly detection in high-dimensional embedding spaces, unsupervised coordination detection on multi-layer graphs, and agentic AI systems. Finally, this survey addresses challenges posed by GenAI, including the difficulty of tracking rapidly changing threats and multi-level distributional drift, and it outlines a future research agenda focused on detecting anomalous clusters and building anticipatory and resilient systems. This survey provides a comprehensive, lifecycle-based review of methods for the proactive detection of emerging synthetic threats for more resilient information ecosystems.
\end{abstract}


\titlepgskip=-15pt

\let\year\ieeeYear  
 \maketitle
\let\year\realTexYear 

\section{Introduction}
\label{sec:introduction}
\PARstart{R}{esearch} on digital influence operations has historically been reactive, focusing on detection after the spread of inauthentic narratives. This reactive approach primarily focuses on analyzing content provenance, the linguistic style of the content, network propagation patterns, and the credibility of sources \cite{shu2020fake}. These methods are useful for finding known synthetic artifacts that are already out there, but they don't work for finding fresh generated content. This response-based approach suffers from high latency, as intervention can only occur after a narrative has achieved initial spread, and it suffers at adapting to the rapidly evolving tactics of adversarial actors \cite{aclanthology2025}. As a result, this paradigm is fundamentally unsuitable to address the challenge of fresh inauthentic narratives, where no ground truth has yet been established and for which labeled training data does not exist.

\begin{figure}[t!]
\centering
\resizebox{\columnwidth}{!}{%
\begin{tikzpicture}[font=\sffamily\scriptsize]
\node[align=right] (reactive_label) at (-1.5, 2.5) {Reactive\\Detection};
\draw[->, thick, gray] (0, 1.5) -- (10.5, 1.5);
\node[below, gray] at (10.5, 1.5) {Time};
\node[circle, fill=black!70, inner sep=3pt] (creation) at (0.5, 2.5) {};
\draw[->, gray] (creation) -- ++(1, 0);
\node[below] at (0.5, 1.5) {Creation};
\begin{scope}[shift={(2.5, 2.5)}]
    \foreach \x/\y in {0/0, 0.2/0.2, -0.2/0.1, 0.1/-0.2, -0.15/-0.15}
        \node[circle, fill=black!60, inner sep=1.5pt] at (\x, \y) {};
\end{scope}
\draw[->, gray] (3, 2.5) -- ++(1, 0);
\node[below] at (2.5, 1.5) {Seed};
\begin{scope}[shift={(5, 2.5)}]
    \foreach \angle in {0, 45, ..., 315}
        \node[circle, fill=black!60, inner sep=1.5pt] at (\angle:0.4) {};
    \foreach \angle in {20, 70, ..., 340}
        \node[circle, fill=black!60, inner sep=1.5pt] at (\angle:0.2) {};
    \node[circle, fill=black!60, inner sep=1.5pt] at (0,0) {};
\end{scope}
\draw[->, gray] (5.8, 2.5) -- ++(0.5, 0);
\node[below] at (5, 1.5) {Spread};
\fill[pattern=north east lines, pattern color=red!80] (5.5, 1.5) rectangle (10, 1.9);
\draw[red!80, thick] (5.5, 1.5) rectangle (10, 1.9);
\node[white] at (7.75, 1.7) {\strut Reaction Gap};
\fill[pattern=north east lines, pattern color=red!60] (6.5, 2.4) rectangle (9.5, 2.6);
\draw[red!60, thick] (6.5, 2.4) rectangle (9.5, 2.6);
\node at (8, 3.0) {\tikz{
    \draw[thick] (0,0) circle (3pt);
    \draw[thick] (2pt, -2pt) -- (4pt, -4pt);
}};
\node at (8, 2.8) {Detection};
\node at (10, 3.0) {\tikz{
    \draw[thick] (0,0) circle (4pt);
    \draw[thick] (-3pt, -3pt) -- (3pt, 3pt);
    \draw[thick] (-3pt, 3pt) -- (3pt, -3pt);
}};
\node at (10, 2.8) {Debunking};
\draw[dashed, gray] (6.5, 2.4) -- (6.5, 1.5);
\draw[dashed, gray] (9.5, 2.4) -- (9.5, 1.5);
\draw[dashed, gray] (10, 1.9) -- (10, 2.5);
\node[align=right] (proactive_label) at (-1.5, -0.5) {Proactive\\Anticipation};
\draw[->, thick, gray] (0, -1.5) -- (10.5, -1.5);
\node[below, gray] at (10.5, -1.5) {Time};
\fill[green!40, rounded corners=5pt] (0.5, -1.3) rectangle (8.5, -0.7);
\node at (4.5, -1.0) {Anticipatory Alignment};
\node[below, align=center] at (2.5, -1.6) {Context Monitoring};
\node[below, align=center] at (5.5, -1.6) {Anomaly\\($I$ vs $Z$)};
\node[below, align=center] at (8.5, -1.6) {Pre-bunking /\\Intervention};
\draw[dashed, gray] (0.5, -1.5) -- (0.5, -0.5);
\draw[dashed, gray] (3.5, -1.5) -- (3.5, -0.5);
\draw[dashed, gray] (5.5, -1.5) -- (5.5, -0.5);
\draw[dashed, gray] (8.5, -1.5) -- (8.5, -0.5);
\node at (2.0, -0.3) {$b/l$};
\draw[thick] (2.4, -0.3) -- (2.5, -0.3) -- (2.6, -0.1) -- (2.7, -0.5) -- (2.8, -0.1) -- (2.9, -0.3) -- (3.0, -0.3);
\draw[gray, thin] (4.8, -0.5) -- (4.8, 0.0) node[right, black] {Critical Vulnerability};
\draw[gray, thin] (4.8, -0.5) -- (5.8, -0.5);
\draw[red, thick] (4.8, -0.4) .. controls (5.2, -0.4) .. (5.5, 0.0) node[left] {$I$};
\draw[blue, thick] (4.8, -0.45) -- (5.5, -0.45) node[below] {$Z$};
\draw[->, red, thick] (5.8, 0.0) -- (5.5, -0.1); 
\node at (7, -0.3) {\tikz{
    \fill[black] (0,0) -- (0.15, 0) -- (0.15, -0.15) .. controls (0.075, -0.3) .. (0, -0.15) -- cycle;
    \fill[black] (0,0) -- (-0.15, 0) -- (-0.15, -0.15) .. controls (-0.075, -0.3) .. (0, -0.15) -- cycle;
    \draw[white, thick] (0, -0.05) -- (0, -0.2);
    \draw[white, thick] (-0.1, -0.1) -- (0.1, -0.1);
}};
\end{tikzpicture}
}
\caption{Comparison between reactive and proactive detection.\label{fig1}}
\end{figure}
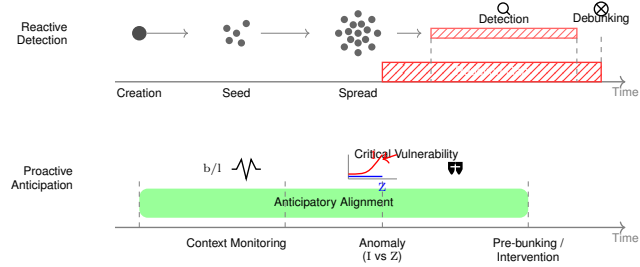

The advent of Transformers and Generative Adversarial Networks (GANs) has rendered this chronic challenge more urgent. GenAI has fundamentally accelerated the creation of content with much lower cost, enabling malicious actors to generate targeted content with high quality at a very low cost. This creates a moving target swarm, where a vast volume of semantically related but stylistically unique pieces of content overwhelm the traditional way of defenses. Since the rate of synthetic content generation now enormously outpaces manual auditing \cite{pnas2024}, relying on static ground truth has become a bottleneck. Many studies confirm that AI systems can ``process thousands of claims in a matter of seconds,'' whereas ``human claim assessments require hours or days'' \cite{fhss2025}.

In response to these systematic bottlenecks, a body of research has emerged around proactive defenses \cite{mdpi2024}. However, much existing survey literature adopts a model-centric view, focusing on methods to strengthen Large Language Models (LLMs) internally. However, an effective proactive approach must be ecosystem-aware, understanding adversarial information flows as a dynamic, socio-technical process instead of a model failure. This survey aims to bridge the gap between reactive ecosystem monitoring and proactive model hardening by synthesizing current research on proactive ecosystem anticipation using signals from the information environment to foresee emerging narratives. This survey identifies and categorizes the methodological limitations of the reactive, content-centric paradigm, contrasting them with emerging proactive, lifecycle-based approaches. The main goal of this work is to combine socio-technical lifecycle models with proactive computational methods.

\section{The Inauthentic Narrative Lifecycle: A Taxonomy for Detection}
Current literature indicates that effective forecasting and mitigation requires frameworks that perceive inauthentic narratives not as static artifacts, but as dynamic, multi-faceted processes. To achieve this, we need process-oriented frameworks to acquire a comprehensive understanding of the complete lifecycle of inauthentic narratives that helps us identify essential intervention points \cite{commtheory2025}. These frameworks include significant feedback loops that consider how synthetic narratives evolve and adapt in response to countermeasures and changing public sentiment \cite{tandf2024}.

\subsection{Theoretical Foundation: From Linear Transmission to Circular Interaction}
To structure our analysis of proactive detection, we revisit the foundational functionalist model of communication proposed by Lasswell (1948). Lasswell organized communication analysis around five main dimensions: ``Who (1) says What (2) in Which channel (3) to Whom (4) with What effect (5)?'' \cite{lasswell1948}. This framework provides the essential components of information flow, but it depicts these elements as a linear transmission sequence. However, in the era of Generative AI and algorithmic social media, adversarial content does not flow linearly. It propagates through recursive feedback loops. To address this socio-technical complexity while retaining the clarity of foundational theory, we adopt the C5 Interaction Model. As illustrated in Table \ref{tab1}, the C5 framework can be understood as a socio-technical adaptation of Lasswell's 5W model, specifically re-engineered to highlight the precursors of inauthentic campaigns essential for proactive detection.

\begin{table*}[t]
\caption{Mapping Lasswell's 5W Model to the C5 Interaction Framework}
\label{tab1}
\setlength{\tabcolsep}{4pt}
\renewcommand{\arraystretch}{1.3}
\begin{tabularx}{\textwidth}{|p{3.5cm}|p{4cm}|X|}
\hline
\textbf{Lasswell's 5W (1948)} & \textbf{C5 Component} & \textbf{Description \& Proactive Relevance} \\
\hline
To Whom (Audience) & 1. Context & The pre-existing social vulnerabilities (e.g., polarization) that make the ``Whom'' susceptible. Monitoring this allows anticipating where narratives will take root. \\
\hline
Who (Communicator) & 2. Causes & The actors (state, for-profit) and their intent. Detecting ``Who'' (e.g., via coordination detection) allows intervention before content spreads. \\
\hline
Says What (Message) & 3. Content & The artifact itself (text, image). Traditional detection focuses here, often too late. \\
\hline
In Which Channel & 4. Cycle of Amplification & The algorithmic and social mechanisms of spread. GenAI accelerates this channel efficiency. \\
\hline
With What Effect & 5. Consequences & The societal impact. Feedback from consequences alters the Context for the next cycle. \\
\hline
\end{tabularx}
\end{table*}

By mapping the ecosystem this way, the C5 model shifts the focus from merely analyzing the ``Message'' (Content) to monitoring the ``Who'' (Causes) and ``To Whom'' (Context)—the two stages that exist before a viral narrative fully emerges \cite{springer2025}.

\subsection{Adopting the C5 Interaction Model for Survey Organization}
With the theoretical lineage established, we now operationalize the framework. Unlike static transmission models, the C5 model focuses on the dynamic feedback loops between five core components: Context, Causes, Content, Cycle of Amplification, and Consequences. These components function as an interrelated ecosystem where the consequences of one narrative reshape the context for the next, creating a continuous cycle. This survey organizes the analysis of emerging inauthentic narratives and proactive detection methodologies according to these five stages, moving from the latent precursors to the kinetic spread \cite{springer2025}.

The \textbf{Context} is the first step in the lifecycle. It is the fertile ground in which synthetic narratives are seeded and grow. This includes the broader social and technological landscape. High levels of information entropy and ecosystem instability render audiences susceptible to adversarial manipulation \cite{masscomm2023, sage2022}. To assess current capabilities in moving beyond qualitative description, this survey reviews techniques for quantifying 'Context' mathematically. As detailed in Section V, we operationalize societal vulnerability using the parameters $b$ (skeptic influence) and $l$ (skeptic recruitment) from the SEIZ epidemiological model, treating the degradation of these parameters as a quantifiable signal of a weakened information ecosystem \cite{acm2013, springerchapter2024}.

The second step, \textbf{Causes}, looks at the adversarial actors and their motivations. This domain is driven by a diverse array of entities, ranging from sophisticated state-sponsored operations and for-profit networks to ideologically driven groups and individual trolls seeking social validation \cite{tandf2024}.

These actors generate \textbf{Content}, the specific media artifact—such as a textual claim, a deepfake video, or a viral meme—that carries the deceptive message. Historically, reactive detection efforts have myopically focused on this single stage, analyzing linguistic style and authenticity to determine correctness \cite{brainsci2023}.

However, content alone is useless without the \textbf{Cycle of Amplification}, the complicated process by which a narrative becomes popular. This stage is driven by a convergence of technical and human factors. Algorithmic amplification on digital platforms accelerates spread \cite{acm2023, atlantic2025}, while cognitive biases and social network structures make viral cascades easier \cite{nber2021, thaijo2025, springer2025b}.

Finally, the lifecycle concludes with \textbf{Consequences}, which refer to the tangible impacts on both individuals and society. These range from the breakdown of democratic norms and increased social division to acute public health emergencies and, in extreme cases, the incitement of real-world violence \cite{springer2025}.

\subsection{A Scaffold for Interdisciplinary Synthesis}
The C5 model serves as a robust socio-technical scaffold that enables the integration of different research streams. Computational techniques for detecting Coordinated Inauthentic Behavior is useful for detecting inauthentic narratives in the Causes stage, while network science is crucial for understanding the Cycle of Amplification \cite{springer2025}. This framework functions as a prescriptive diagnostic tool. A campaign driven by sophisticated adverse actors (``Causes'') requires different responses (e.g., account takedowns) compared to information that spreads organically due to a resonant narrative (``Content'') exploiting pre-existing anxieties (``Context'').

The non-linear nature of the C5 model holds profound implications for proactive detection. The elements support each other, creating a feedback loop in which ``Consequences'' change the ``Context'' for the next cycles \cite{springer2025}. This means one can monitor the ``Context'' for signals of rising societal disagreement or the ``Causes'' for behavioral indicators of actor mobilization, using these as precursors for an impending information threat.

\subsection{A Critical Comparison of Lifecycle Frameworks for Proactive Detection}
The selection of the C5 model is based on its better suitability for a proactive paradigm compared to alternative frameworks. The C5 model's primary strength for anticipation lies in its non-linear, socio-technical structure, which elevates ``Context'' and ``Causes'' to primary, actionable elements rather than mere background information. The analysis of the C5 model confirms that its key insight is that the lifecycle route is predominantly influenced by the social context, which makes it possible to identify dangerous interactions before a campaign enters ``Content'' stage \cite{springer2025}.

This directly enables proactive interventions like ``pre-bunking,'' a strategy that directly targets and modifies the ``Context'' element to immunize a population against future narratives. Empirical studies have validated the use of such pre-emptive, source-focused inoculation as a countermeasure, though its effectiveness varies \cite{pnas2025}. This proactive, precursor-focused approach contrasts sharply with alternative frameworks:

\begin{itemize}
    \item \textbf{The Reactive Operational Pipeline:} Traditional operational pipelines (e.g., Detect $\rightarrow$ Verify $\rightarrow$ Debunk) are inherently reactive. They function only after the Content exists. C5, by emphasizing Context and Causes, provides computationally tractable precursors that exist before the Content is generated.
    \item \textbf{The SIR Model:} While epidemiological models (SIR) describe spread, they often lack the socio-technical nuance of why spread occurs \cite{physica2011}. C5 provides the semantic layer (Context/Causes) that explains the parameters of the mathematical propagation models.
\end{itemize}

The C5 model, as adopted in this survey, acts as the scaffold that organizes both. It provides a structure (Context, Causes, Content...) where the computational tasks from the pipeline model can be applied as tools and the theoretical dynamics from the process model can be understood and predicted \cite{springer2025}. It is this synthesis that makes the C5 model the key to unlocking a proactive paradigm, as it is the only framework that provides computationally tractable precursors (``Context'' and ``Causes'') that can be monitored before ``Content'' (the inauthentic narrative) emerges.

\section{The Creation and Seeding of Fresh Narratives (Causes \& Content)}
In this analysis, we categorize all non-factual or manipulated information flows under the broader umbrella of Synthetic Noise. We distinguish between two distinct vectors of this noise based on intent:

\begin{itemize}
    \item \textbf{Adversarial Inauthenticity:} Intentional campaigns driven by Coordinated Inauthentic Behavior (CIB) and malicious actors.
    \item \textbf{Stochastic Falsehoods:} Unintentional false artifacts arising from model hallucinations and probabilistic errors.
\end{itemize}

While their origins differ---one malicious, one mechanical---their downstream effect on ecosystem resilience is functionally identical: both degrade the signal-to-noise ratio of the information environment. Therefore, this survey treats both as structural threats to information integrity that require proactive anticipation, regardless of the creator's intent.

\subsection{Cause}
The cause of the creation of malicious content depends on the creators and their motives to spread synthetic narratives.

\subsubsection{Democratization of creation}
The advent of GenAI LLMs has lowered the time and cost for content creation. ``Generative models, especially deep learning based generative models such as GANs, Variational Autoencoders (VAEs), autoregressive models, and diffusion-based models have demonstrated remarkable success in generating realistic and diverse content in various domains \cite{arxiv2024}. Unlike traditional methods that require manual effort and expertise, generative AI can rapidly generate content at scale, saving valuable time and resources for content creators. This efficiency is particularly valuable in fast-paced industries where content needs to be produced quickly and frequently to stay relevant \cite{efai2024}. It is now not too difficult to automate video, image, and text creation that looks very realistic at virtually no cost. This aids the creation of personalized targeted content further increasing challenges in news, media, and information integrity \cite{arxiv2024b}. Due to the increasing ease of creation, the number of creators will tend to increase, commensurately increasing the number of malicious actors.

\subsubsection{Creation of content on an industrialized scale}
The automated scalability of GenAI allows for the creation of content at an unprecedented scale. There are already several instances of deepfakes being used for financial fraud. For instance, on February 2, 2024, a finance worker was tricked into paying out \$25 million in transfers to criminals using Deep Fake video \cite{easttom2025}. Producing content at such a scale also reduces overhead costs by minimizing the reliance on extensive human labor for content generation.

\subsubsection{Stochastic Falsehoods: Unintentional Instability from AI Hallucinations}
While many causes would be due to malicious intent of content creators, sometimes an honest attempt to create meaningful content can lead to seeding of non-factual artifacts. This may happen if the model used to produce the said content hallucinates. These can occur due to biases in training data, limited context, ambiguous prompts, adversarial attacks, overfitting, and model architecture \cite{reddy2024}. This non deliberate spread of generation of hallucinated artifacts can have significant practical implications, such as instances where a model hallucinates while researching critical decisions while treating an ailment, or loss of money due to a artifact-based decision while financial trading. Unlike adversarial inauthenticity, these artifacts are not designed to deceive. However, when amplified by algorithmic systems, they function as 'misinformation without intent,' saturating the ecosystem with plausible but false narratives.

\subsection{Content}
The nature of the content through which synthetic narratives propagates has also been affected by Gen AI. Some of the ways this has changed are:

\subsubsection{Hyper-realistic content}
It is becoming increasingly difficult to identify AI generated vs human created content. For instance, AI-generated faces are now widely available (e.g., this-person-does-not-exist.com) and are being used for both prosocial and nefarious purposes, from finding missing children \cite{chandaliya2022} to transmitting inauthentic political narratives via fake user accounts \cite{hatmaker2020} \cite{miller2023}. As a result, easily discernible manipulations are becoming less prevalent, while synthetic content that is nearly indistinguishable from authentic sources is becoming increasingly common.

\subsubsection{Hyper personalized and targeted content}
By leveraging Gen AI, it is much easier to create content targeted towards an individual. Synthetic content spread is no longer limited to just broadcasting false information. Moreover, the speed with which we can generate content has been increasing with time. It is now possible to quickly mould synthetic content to align with a consumer’s online preferences and then distribute across digital platforms. A study by Bedi et al \cite{bedi2024} reported 30\% increase in click through rate when GPT based models were used for personalized advertisements. Patil et al \cite{patil2025} has listed a review on such research on AI driven hyper personalization and has proposed a theoretically validated approach to address complexities of digital ecosystem dynamics.

\subsubsection{Varied content (or any other type of content)}
Its quite easy to create a plethora of content using gen ai without needing to be expert in all the topics the content is based on. Suri et al \cite{suri2024} has mentioned how Gen AI is utilized to create content with an accuracy not too far from manual creation. Jacob et al \cite{jacob2023} has created a comprehensive survey on AI content generator and shown how these can create quality content while saving time, expenses, and enhancing quality. There has also been a lot of research on how AI can assist in creating content in fields like education \cite{blagoev2023, faccia2023}, marketing \cite{grewal2025, aldous2024}, and social media content \cite{sancanin2022, aldous2024}.

\subsection{How does this amount of synthetic content spread?}
One emerging way of spreading inauthentic narratives is using a methodology called Coordinated Inauthentic Behavior (CIB). It is a manipulative communication tactic that uses a mix of authentic, fake, and duplicated social media accounts to operate as an adversarial network (AN) across multiple social media platforms \cite{murero2023}. While not inherently malicious, such coordination becomes problematic when it is deceptive in nature, e.g., when inauthentic accounts present themselves as genuine individuals or manipulate visibility through algorithmic gaming \cite{arxiv2025}. For detecting such spread of synthetic artifacts rather than auditing the veracity of a claim, it is more important to analyze the pattern of dissipation of such content and evaluate if that seems organic. There are several ways in which we can detect such spread of information. One interesting method is to use Bayesian inference to identify groups of accounts that share similar account-level characteristics and target similar narratives \cite{smith2024}.

\subsection{Role of Gen AI and CIB in Inauthentic Narrative lifecycle}
As with creation of content, Gen AI also enhances the effectiveness of CIB. They share a symbiotic relationship. Pointing to the amount of varied content Gen AI can produce, it can aid in creating very diverse content for the same inauthentic narrative through the large number of bots CIB employs. This can pose challenges to signature based detection of such content.

On the other hand, CIB solves the scaling problem for AI created synthetic artifacts. It acts as a means to introduce such artifacts to digital platforms at such a scale that the algorithms can begin recognizing and incorporating such content into their recommendations. This further amplifies the distribution of such content.

Early techniques of CIB detection involved examining the profile of the bot to find inaccuracies in the metadata. More recently, bad actors employing CIB have started using compromised accounts that might be aged. We ideally want to shift our focus from detecting fake accounts to fake behavior. Grimme et al. (2023) proposed a way to detect digital fingerprints of such propagation of content. They employed a time-series-based interpretation of topic-based campaign detection mechanism. This approach works well for large bursts of AI content to ensure that detection quality surpasses false positive rate of their approaches \cite{grimme2023}.

\section{The Propagation of Fresh Narratives (Cycle of Amplification)}
Once a fresh narrative is created and seeded, its potential impact is determined by its propagation through the Cycle of Amplification, a stage based on a convergence of human psychology, network structures, and platform architecture \cite{springer2025}. This spread is heavily controlled by platform recommendation engines, which is also called Algorithmic Amplification \cite{policyreview2023, acm2021}. These systems, optimized for specific business goals such as engagement which sometimes can have trade-offs with information accuracy, act as an external forcing function that dynamically speeds up the spread of sensational content. These algorithms effectively change the parameters of transmission—such as the contact rate ($\beta$) in epidemiological terms—allowing inauthentic narratives to achieve viral saturation faster than organic verification can occur by prioritizing emotionally charged narratives that trigger confirmation bias or in-group signaling \cite{ieee2023, springer2025a}. To understand these accelerated dynamics, we can model the pattern of narrative spread using advanced, mathematical frameworks: epidemiological models for macro-level belief states and Hawkes processes for micro-level event cascades.

\subsection{Modeling Information Spread as a Contagion: Epidemiological Models}
One of the prominent approaches is to treat information propagation as a biological contagion, adapting models from epidemiology to describe the flow of signals through a population \cite{springerchapter2024}. These models split a population into multiple states and use ordinary differential equations (ODEs) to model the transitions between them. The classic SIR (Susceptible, Infected, Recovered) model provides a foundational analogy: people are initially Susceptible to a piece of synthetic narrative, then they become Amplifiers when they adopt and share it, and finally Recover, either by becoming immune (e.g., through debunking) or by losing interest \cite{physica2011}.

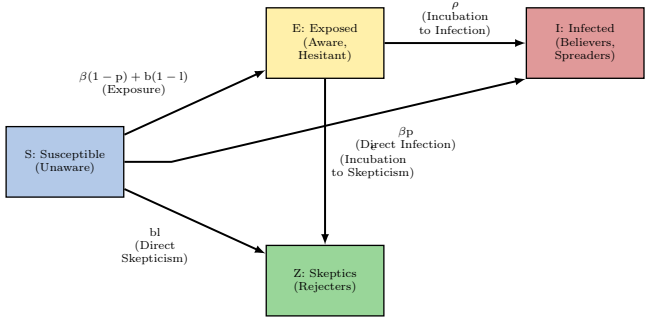
\begin{figure}[t!]
\centering
\resizebox{\columnwidth}{!}{%
\begin{tikzpicture}[
    node distance=2cm, 
    auto,
    font=\sffamily\scriptsize,
    box/.style={rectangle, draw, thick, minimum width=2.5cm, minimum height=1.5cm, align=center, font=\bfseries\scriptsize},
    arrow/.style={->, >=latex, very thick},
    lbl/.style={align=center, font=\scriptsize}
]
    \node[box, fill=teal!60!blue!30] (S) {S: Susceptible\\(Unaware)};
    \node[box, fill=yellow!80!orange!40, above right=1cm and 3cm of S] (E) {E: Exposed\\(Aware,\\Hesitant)};
    \node[box, fill=green!60!black!40, below right=1cm and 3cm of S] (Z) {Z: Skeptics\\(Rejecters)};
    \node[box, fill=red!70!black!40, right=3cm of E] (I) {I: Infected\\(Believers,\\Spreaders)};
    \draw[arrow] (S) -- node[lbl, above left] {$\beta(1-p) + b(1-l)$ \\ (Exposure)} (E);
    \draw[arrow] (S) -- node[lbl, below left] {$bl$ \\ (Direct \\ Skepticism)} (Z);
    \draw[arrow] (S.east) -- ++(1,0) -- node[lbl, pos=0.5, below right] {$\beta p$ \\ (Direct Infection)} (I.south west);
    \draw[arrow] (E) -- node[lbl, above] {$\rho$ \\ (Incubation\\to Infection)} (I);
    \draw[arrow] (E) -- node[lbl, right] {$\epsilon$ \\ (Incubation\\to Skepticism)} (Z);
\end{tikzpicture}
}
\caption{SEIZ Model. In a proactive defense framework, interventions such as pre-bunking are mathematically modeled as an external force increasing the transition rate $b$ and $l$, effectively accelerating the flow from Susceptible ($S$) to Skeptics ($Z$) before infection occurs.\label{fig3}}
\end{figure}

More nuanced variants like the SEIZ (Susceptible, Exposed, Infected, Skeptics) model, on the other hand, offer a more robust framework for this domain \cite{springer2025b}. The SEIZ model introduces two critical additional compartments:
\begin{itemize}
    \item \textbf{Exposed (E):} This accounts for a group of users that has not yet determined whether to propagate the inauthentic narrative after they have encountered it \cite{springer2025b}.
    \item \textbf{Skeptics (Z):} This group of people is exposed to the narrative but actively chooses to not spread it \cite{springer2025b}.
\end{itemize}

The dynamics of the SEIZ model can be represented by the following system of ODEs:
\begin{align*}
dS/dt &= -(\beta \cdot S \cdot I)/N - (b \cdot S \cdot Z)/N \\
\begin{split}
dE/dt &= (1-p) \cdot (\beta \cdot S \cdot I)/N + (1-l) \cdot (b \cdot S \cdot Z)/N \\
      &\quad - (\rho \cdot E \cdot I)/N - \epsilon \cdot E
\end{split} \\
dI/dt &= p \cdot (\beta \cdot S \cdot I)/N + (\rho \cdot E \cdot I)/N + \epsilon \cdot E \\
dZ/dt &= (l \cdot b \cdot S \cdot Z)/N
\end{align*}

Where:
\begin{itemize}
\item $S, E, I (Amplifiers), Z$ represent the absolute count of individuals in each compartment. The transmission terms are normalized by the total population $N$ to model frequency-dependent social contact dynamics.
\item $\beta$: Contact rate between Susceptible ($S$) and Infected ($I$) nodes.
\item $b$: Contact rate between Susceptible ($S$) and Skeptics ($Z$) nodes.
\item $\rho$: Contact rate between Exposed ($E$) and Infected ($I$) nodes.
\item $\epsilon$: Incubation rate.
\item $p$: Probability of transitioning from $S \to I$ given contact with adopters.
\item $1-p$: Probability of transitioning from $S \to E$ given contact with adopters.
\item $l$: Probability of transitioning from $S \to Z$ given contact with skeptics.
\item $1-l$: Probability of transitioning from $S \to E$ given contact with skeptics.
\end{itemize}

This structure allows us to mathematically define the virality of resilience—how effectively the organic baseline or doubt is competing with the synthetic noise \cite{wandb2023}. The inclusion of the $Z$ (Skeptic) compartment allows this model to account for counter-narratives or debunking efforts. A proactive defense system can estimate the parameter $b$ (influence of skeptics) to determine if the network has a natural immune response or if intervention is required. If $I(t)$ (Infected) is growing exponentially while $Z(t)$ (Skeptics) remains flat, it means that the narrative has hit a critical vulnerability in the Context. Comparative studies have shown that the SEIZ model fits real-world social media data better compared to simpler SIR models. This shows how important it is to model both deliberation and skepticism in the information diffusion process \cite{springer2025a}.

\subsection{Modeling Burstiness and Cascades: Hawkes Processes}
Epidemiological models can capture macro-level dynamics, but they struggle to represent the bursty, cascading nature of online sharing. Hawkes processes, a class of self-exciting point processes, offers a more robust framework for this \cite{frontiers2022, acl2016}. The main idea is that when an event (e.g., a retweet) happens, it temporarily increases the probability of future events. This is captured by the conditional intensity function, $\lambda(t)$, which defines the instantaneous probability of an event at time $t$:
\begin{equation}
\lambda(t) = \mu(t) + \sum_{t_i < t} \phi(t - t_i)
\end{equation}
where $\mu(t)$ is the background (exogenous) rate, and the summation represents the self-exciting (endogenous) component from past events $t_i$, governed by a kernel function, $\phi(s)$, that models the decaying influence over time \cite{frontiers2022}. Common choices for the kernel include the exponential kernel $\phi(t) = \alpha e^{-\beta t}$, representing a rapid decay of influence.

In the context of GenAI-driven botnets, the background rate $\mu$ is no longer constant or purely organic. A coordinated botnet can artificially inflate $\mu$, simulating widespread organic interest (astroturfing) \cite{acm2025}. Proactive detection algorithms must effectively disentangle the self-exciting component (viral spread) from the exogenous component (coordinated seeding) to identify inauthentic amplification. If $\mu$ spikes simultaneously across uncorrelated user clusters, it is a strong signature of Coordinated Inauthentic Behavior (CIB) acting on the ``Causes'' level of the C5 model.

Advanced applications of Hawkes processes such as Two-stage models can capture the entire lifecycle of a inauthentic narrative, from when it starts to its eventual correction. In two-stage models, the first phase of information spread is followed by a second phase, starting at a correction time ($t_c$), where the spread of debunking information acts as its own cascade and overtakes the original information \cite{plos2021b}. This approach is powerful because it not only helps predict a narrative's trajectory but also offers a quantitative method for inferring the precise moment a narrative's credibility begins to collapse within a population.

\subsection{Proactive Detection via Propagation Modeling: Operationalizing C5}
By linking these mathematical frameworks to the C5 Interaction Model, we can derive actionable signals for proactive detection, identifying threats based on their behavior rather than just their content. The inclusion of the Skeptic ($Z$) compartment allows the SEIZ model to mathematically operationalize the C5 Interaction Model, specifically linking the kinetic dynamics of the Cycle of Amplification to the latent stability of the Context. Unlike traditional infection models, SEIZ accounts for counter-narratives or debunking efforts through parameters $b$ (the rate of contact with skeptics) and $l$ (the probability of persuasion upon contact), which serve as real-time, quantifiable proxies for the Context's resilience \cite{acm2013, springerchapter2024}. A proactive defense system can continuously estimate these parameters to assess the network's natural immune response. If the Infected ($I$) compartment grows exponentially while the Skeptics ($Z$) compartment remains flat, it signals a collapse in $b$ (reach) or $l$ (trust), indicating that the narrative has exploited a critical vulnerability in the Context—such as deep polarization or eroding institutional trust—where the verified information is either not reaching or not convincing the susceptible population \cite{acm2013}.

While SEIZ models macro-level belief states, Hawkes Processes offer a granular, event-driven detection mechanism for the Causes and Cycle of Amplification components of the C5 model. By mathematically decomposing the intensity of information flow into a background rate ($\mu$) and a self-exciting component, this framework allows defense systems to disentangle organic viral spread (Amplification) from artificial seeding operations (Causes) \cite{ieee2019}. A synchronized spike in $\mu(t)$ across uncorrelated user nodes suggests that a narrative is being externally forced into the system (e.g., by a botnet), rather than spreading through genuine social contagion \cite{physreve2017}.

\begin{table*}[t]
\caption{Comparative Analysis of Propagation Modeling Paradigms}
\label{tab2}
\setlength{\tabcolsep}{4pt}
\renewcommand{\arraystretch}{1.3}
\begin{tabularx}{\textwidth}{|p{3cm}|X|X|}
\hline
\textbf{Feature} & \textbf{Epidemiological Models} & \textbf{Hawkes Processes} \\
\hline
Core Analogy & Spread of biological disease in a population. & Self-exciting cascade of events (e.g., aftershocks following an earthquake). \\
\hline
Unit of Analysis & Population compartments (e.g., number of Susceptible, Infected individuals). & Discrete events in continuous time (e.g., timestamps of individual shares). \\
\hline
Key Model Variants & SIR (Susceptible, Infected, Recovered), SEIZ (Susceptible, Exposed, Infected, Skeptics). & Time-Dependent Hawkes (TiDeH), Two-Stage Correction Models. \\
\hline
Mathematical Basis & Systems of Ordinary Differential Equations (ODEs). & Stochastic point processes with a conditional intensity function. \\
\hline
Core Assumptions & Homogeneous mixing within populations; transitions between discrete states. & Past events trigger future events; influence decays over time. \\
\hline
Strengths & Intuitive, captures macro-level belief dynamics, can incorporate psychological states like skepticism. & Captures the bursty, cascading nature of online sharing, models temporal clustering well. \\
\hline
Limitations & Often ignores network structure and individual heterogeneity; poor at modeling burstiness. & Can be computationally intensive; less intuitive for modeling population-level belief states. \\
\hline
\end{tabularx}
\end{table*}

\section{Proactive Detection Methodologies}

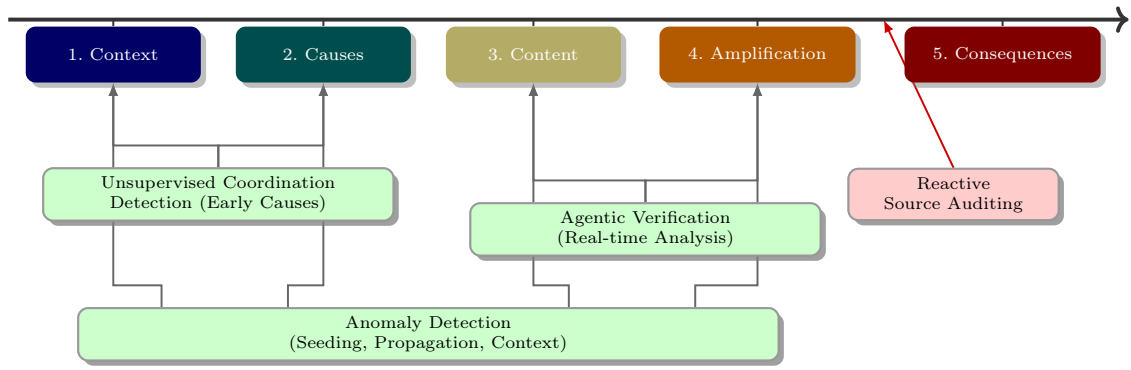
\begin{figure*}[t!]
\centering
\resizebox{0.85\textwidth}{!}{
\begin{tikzpicture}[
    font=\sffamily\scriptsize,
    box/.style={rounded corners, fill=#1, text=white, align=center, minimum height=0.8cm, drop shadow, font=\bfseries\scriptsize},
    detect/.style={rectangle, rounded corners, draw=black!40, thick, fill=#1, align=center, drop shadow, font=\scriptsize},
    arrow/.style={->, >=latex, thick, color=gray!80!black}
]
\draw[->, ultra thick, black!80] (0,0.5) -- (16,0.5);
\node[box=blue!40!black, minimum width=2.5cm] (c1) at (1.5,0) {1. Context};
\node[box=teal!60!black, minimum width=2.5cm] (c2) at (4.5,0) {2. Causes};
\node[box=yellow!60!black!90, text=white, minimum width=2.5cm] (c3) at (7.5,0) {3. Content};
\node[box=orange!70!black, minimum width=2.8cm] (c4) at (10.7,0) {4. Amplification};
\node[box=red!50!black, minimum width=2.8cm] (c5) at (14.2,0) {5. Consequences};
\foreach \n in {c1,c2,c3,c4,c5} \draw[thick, black!80] (\n.north) -- (\n.north |- 0,0.5);

\path (3, -2) node[detect=green!20!white, minimum width=5cm] (unsupervised) {Unsupervised Coordination\\Detection (Early Causes)};
\path (9.1, -2.5) node[detect=green!20!white, minimum width=5cm] (agentic) {Agentic Verification\\(Real-time Analysis)};
\path (6, -4) node[detect=green!20!white, minimum width=10cm] (anomaly) {Anomaly Detection\\(Seeding, Propagation, Context)};
\path (13.5, -2) node[detect=red!20!white, minimum width=3cm] (reactive) {Reactive\\Fact-checking};

\begin{pgfonlayer}{background}
    \coordinate (bus_u) at (0, -1.3);
    \coordinate (bus_a) at (0, -1.8);
    \coordinate (bus_an) at (0, -3.3);

    \draw[arrow] (unsupervised.north) -- (unsupervised.north |- bus_u) -| (c1.south);
    \draw[arrow] (unsupervised.north) -- (unsupervised.north |- bus_u) -| (c2.south);

    \draw[arrow] (agentic.north) -- (agentic.north |- bus_a) -| (c3.south);
    \draw[arrow] (agentic.north) -- (agentic.north |- bus_a) -| (c4.south);

    \path (anomaly.north west) -- (anomaly.north east)
        coordinate[pos=0.12] (an1)
        coordinate[pos=0.3] (an2)
        coordinate[pos=0.7] (an3)
        coordinate[pos=0.88] (an4);

    \draw[arrow] (an1) -- (an1 |- bus_an) -| (c1.south);
    \draw[arrow] (an2) -- (an2 |- bus_an) -| (c2.south);
    \draw[arrow] (an3) -- (an3 |- bus_an) -| (c3.south);
    \draw[arrow] (an4) -- (an4 |- bus_an) -| (c4.south);

    \draw[arrow, red!80!black] (reactive.north) -- (12.5, 0.5); 
\end{pgfonlayer}

\node[detect=green!20!white, minimum width=5cm] at (3, -2) {Unsupervised Coordination\\Detection (Early Causes)};
\node[detect=green!20!white, minimum width=5cm] at (9.1, -2.5) {Agentic Verification\\(Real-time Analysis)};
\node[detect=green!20!white, minimum width=10cm] at (6, -4) {Anomaly Detection\\(Seeding, Propagation, Context)};
\node[detect=red!20!white, minimum width=3cm] at (13.5, -2) {Reactive\\Source Auditing};

\end{tikzpicture}
}
\caption{Inauthentic Narrative Lifecycle based on C5 Model and detection stages.\label{fig4}}
\end{figure*}

This section surveys three complementary computational approaches for proactive detection. In the following analysis, we distinguish between empirically validated techniques (such as anomaly detection) and nascent, experimental proposals (such as agentic verification) to provide a clear assessment of current technological maturity.

\subsection{Defining the Target: The "Emerging Narrative"}
Proactive detection is aimed at identifying an emerging narrative in its nascent stages. This extends beyond a new topic; it can be a novel framing of an existing event, a story arising to fill an information void \cite{pmc2021}, or a subtle manipulation of facts. A computationally tractable proxy for this is a sudden, localized spike in societal divergence around a specific topic, detectable through stance and signal analysis.

\subsection{Approaches based on Multi-Layered Anomaly Detection}
One way to identify emerging inauthentic narratives is to treat them as statistical outliers from an established baseline of normal information flow \cite{plos2025, mdpi2025}. While traditional anomaly detection is often criticized for being reactive, we argue that by integrating the mathematical frameworks of SEIZ and Hawkes processes, this methodology serves as a powerful early warning system \cite{ieee2022}. Proactive systems identify threats in their early stages by shifting the focus from merely detecting anomalous content to detecting anomalous dynamics and context shifts. We categorize these anomalies into three distinct layers:

\subsubsection{Semantic and Topic Anomalies (The Content Layer)}
Emerging narratives appear as outliers in the semantic space at the most granular level. This approach builds a model of normalcy for certain communities or discourse streams, flagging items that deviate significantly from established patterns \cite{mdpi2025}.
\begin{itemize}
    \item \textbf{High-Dimensional Embedding Outliers:} Leveraging the rich semantic representations from transformer-based encoders (e.g., BERT), we can identify emerging narratives as vectors located at a significant distance from the cluster centroids of established discourse in a high-dimensional space ($V \subset \mathbb{R}^d$) \cite{plos2025b}.
    \item \textbf{Topic Burstiness:} A sudden, unexpected spike in the discussion volume of a niche topic or a new combination of keywords often signals the seeding phase of a campaign. This is treated as a burst detection problem, where the anomaly lies in the acceleration of topic frequency rather than the topic itself \cite{aaai2020, acm2002, acm2024}.
\end{itemize}

\subsubsection{Propagation Dynamics Anomalies (The Amplification Layer)}
Moving beyond content, we can leverage Hawkes Processes to detect anomalies in how information spreads. As discussed in Section III, organic viral spread is typically self-exciting, whereas coordinated campaigns often rely on external injection. By mathematically decomposing the intensity function $\lambda(t)$ of a cascade, we can isolate the background rate $\mu(t)$ from the self-exciting component \cite{ieee2019, physreve2017}. A synchronized spike in $\mu(t)$ across uncorrelated user nodes suggests that a narrative is being externally forced into the system (e.g., by a botnet), rather than spreading through genuine social contagion \cite{ieee2019}.

\subsubsection{Vulnerability and Context Anomalies (The Context Layer)}
Recent studies demonstrate the utility of the SEIZ epidemiological model to detect anomalies in the 'Context,' such as the breakdown of a network's resilience. This represents the most proactive layer of detection.
\begin{itemize}
    \item \textbf{Immune System Failure Monitoring:} Defense systems can monitor the skepticism parameter $b$ (the rate of contact with skeptics) by continuously estimating the parameters of the SEIZ model on live data streams. A statistically significant drop in $b$ signals that the community's natural resistance to manipulation is degrading \cite{acm2013, springerchapter2024, pmc2022}.
    \item \textbf{Latent Vulnerability Detection:} Unlike content detection, which flags specific narratives, this method identifies silent periods where the Exposed ($E$) compartment grows disproportionately to the Skeptics ($Z$). This indicates a critical vulnerability in the social context, warning of a potential outbreak condition before any specific narrative achieves viral saturation \cite{acm2013, springerchapter2024}.
\end{itemize}

\subsection{Using Unsupervised Coordination Detection}
These techniques provide a more proactive signal by identifying the behavior of malicious actors, often before their content is widely understood. The principle is to find groups of accounts exhibiting statistically improbable similarities in their actions—a strong signal of CIB \cite{pmc2023}. Key techniques include:
\begin{itemize}
    \item \textbf{Bayesian Inference:} This statistical method can be used to identify groups of accounts that share similar account-level characteristics (e.g., creation date, profile description) and target similar narratives, even within large and noisy datasets \cite{epj2025}.
    \item \textbf{Fused Networks:} This state-of-the-art approach moves beyond single behavioral indicators. It constructs a multi-layer graph $G=(V, E)$ where edges $E$ are a weighted combination of multiple behavioral similarity matrices (e.g., co-retweet, co-URL sharing, text similarity). Machine learning models, often leveraging node embeddings like node2vec, can then be applied to classify accounts based on their topological properties within this fused network. This method has been shown to achieve significantly higher precision in detecting influence campaigns compared to methods that rely on only a single behavioral trace \cite{acm2025b}.
\end{itemize}

\subsection{Emerging Provenance Frameworks: Agentic AI and Multi-Agent Systems}
While Anomaly and Coordination Detection serve as the ecosystem's smoke detectors—identifying suspicious patterns—they cannot definitively confirm inauthenticity. To bridge the gap between detection and confirmation without the latency of human intervention, the field is turning to Agentic AI. These systems leverage the very technology driving the problem—GenAI—to automate verification at scale. Unlike passive classifiers, these autonomous agents can plan, use external tools, and execute complex reasoning tasks to verify claims flagged by earlier detection stages.

Agentic AI refers to systems, built on LLMs, that can make autonomous decisions, utilize external tools, and take actions to achieve complex goals. Such systems are defined by their ability to evolv[e] beyond passive tools into autonomous agents capable of reasoning, adapting, and acting with minimal human intervention \cite{aisel2025}. This is operationalized through dynamic task decomposition, specialized agent roles, collaborative protocols, and tool integration \cite{neurips2024}.

Standard LLM-based verification (Zero-Shot or Chain-of-Thought) remains prone to hallucination and and often lacks self-corrective mechanisms. A single model often commits to an initial erroneous premise and generates plausible-sounding justifications for it. Agentic AI overcomes this via Dynamic Task Decomposition. Instead of a single inference pass, the verification task is decomposed into sub-routines executed by specialized agents. The architecture is not merely a collaboration but often a structured conflict designed to surface provenance through adversarial testing. We categorize these emerging architectures along a trajectory of increasing cognitive autonomy: from linear pipelines to adversarial dialectics and meta-evaluation systems.

\subsubsection{Sequential Verification: Pipeline Architectures}
Often seen in other research papers, these approaches automate the reactive lifecycle, defining the process as classification, detection, correction, and source identification \cite{icwsm2025}. A leading example proposes a five-agent pipeline (Indexer, Classifier, Extractor, Corrector, Verification), where the core Extractor agent is implemented as a Retrieval-Augmented Generation (RAG) system that queries external knowledge bases \cite{icwsm2025}.

\subsubsection{Adversarial Verification: Multi-Agent Debate (MAD)}
For ambiguous or highly nuanced claims, linear pipelines often fail to capture context. To address this, recent research leverages Multi-Agent Debate (MAD) frameworks, which replace collaboration with structured conflict to surface truth through adversarial testing \cite{emnlp2024a, emnlp2024b}. Early iterations of this approach were vulnerable to debate hacking, where agents prioritized winning the argument over factual accuracy. Current state-of-the-art proposals, such as the TruEDebate (TED) framework, impose rigorous protocols inspired by the Lincoln-Douglas debate format to enforce epistemic discipline \cite{acm2025c}. The TED architecture separates the system into two distinct layers:
\begin{itemize}
    \item \textbf{DebateFlow Agents:} These agents are the core disputants, divided into Proponents ($T_{pro}$) and Opponents ($T_{opp}$). Each agent is assigned a specific stance (Organic/Synthetic) relative to the news item $F$.
    \item \textbf{InsightFlow Agents:} These function as a meta-layer consisting of a Synthesis Agent (which summarizes the state of the debate) and an Analysis Agent (which renders the final verdict).
\end{itemize}
The power of TED lies in its Role-Aware Encoder. The system maintains a Debate Graph where nodes are arguments and edges represent logical relationships (support/attack). The Analysis Agent does not just look at the text; it processes the topology of the debate graph to determine which side's arguments remained unrefuted.

\subsubsection{Meta-Evaluation: The Agent-as-a-Judge (AaaS) Pattern}
This concept is part of a broader, established research field. Some existing literature has systemized the underlying concepts of LLM-as-a-judge \cite{emnlp2025}. A critical body of research has highlighted that the simpler LLM-as-a-Judge (LLMaaS) pattern is foundationally unreliable. These models suffer from cognitive biases, including positional bias, where the quality can be easily hacked by simply altering their order of appearance \cite{acl2024}. This also includes order effects, prompt sensitivity, and other cognitive biases \cite{recsys2025}. This has led academic bodies to warn against using LLM-only evaluation without robust validation from human experts being in the loop \cite{acm2025d}.

A specific, next-generation implementation addressing this, the Agent-as-a-Judge (AaaS) framework, was presented by Zhuge, M., et al. AaaS extends the LLM-as-a-Judge pattern by incorporating agentic features that enable intermediate feedback for the entire task-solving process \cite{icml2025}. Rather than judging only the final output, the AaaS agent can use tools to evaluate another agent's full thought and action trajectory, providing a crucial mechanism for step-by-step reasoning evaluation. On the DevAI benchmark, this AaaS framework achieved ~90\% agreement with human expert evaluations, far exceeding the ~70\% agreement of LLMaaS methods, and did so with a 97\% reduction in time and cost.

\subsubsection{Summary of Strategic Integration}
We can think of these three methodologies as existing on a spectrum of proactivity—that is, how early they can intervene. \textbf{Multi-layered anomaly detection} serves as a rapid early warning system. While it acts upon statistical deviations, it bridges the gap between prediction and reaction by flagging anomalous dynamics and context shifts in their nascent stages, long before viral saturation. \textbf{Unsupervised coordination detection} is a step ahead, offering a more forward-looking approach. It is designed to identify the mobilization of actors as a campaign is first being launched, catching the seed before it becomes a forest fire. \textbf{Agentic AI systems} play a related but distinct role. They are less a single detection method and more a set of powerful tools for real-time verification that can be deployed once a potential narrative has been flagged by one of the other systems. Therefore, current literature suggests that integrating these methodologies offers a more robust defense by covering the full narrative lifecycle. This approach theoretically links the detection of Coordinated Seeding (campaign launch) and Anomalous Propagation (early spread) with Agentic Verification (claim assessment), rather than relying on any single method in isolation. Table \ref{tab3} provides a taxonomy of these proactive methodologies.

\begin{table*}[t]
\caption{Taxonomy of Proactive Detection Methodologies and Implementations}
\label{tab3}
\setlength{\tabcolsep}{4pt}
\renewcommand{\arraystretch}{1.3}
\begin{tabularx}{\textwidth}{|p{2.5cm}|X|X|X|p{2.5cm}|}
\hline
\textbf{Methodology} & \textbf{Core Principle} & \textbf{Key Techniques} & \textbf{Data Requirements} & \textbf{Proactive Signal Type} \\
\hline
Multi-Layered Anomaly Detection & Identify emerging narratives as statistical deviations in content, propagation dynamics, and context stability. & Outlier detection on LLM embeddings, decomposition of Hawkes Processes (intensity function), and continuous estimation of SEIZ model parameters (skepticism $b$). & Baseline data of normal patterns, live interaction streams for parameter estimation. & Content-based / Propagation-based / Context-based \\
\hline
Unsupervised Coordination Detection & Identify the mobilization of inauthentic actors by their coordinated behaviors. & Bayesian inference for group detection, node embeddings on Fused-Networks of behavioral traces. & Social network graphs, user metadata, event logs (posts, shares). & Behavior-based \\
\hline
Agentic AI \& Multi-Agent Systems & Automate real-time claim assessment by using autonomous agents to plan, debate, and judge veracity. & Adversarial Multi-Agent Debate (MAD/TED), Agent-as-a-Judge (AaaS) meta-evaluation, RAG pipelines. & Access to live web search APIs, knowledge bases, and multimodal analysis tools. & Verification-based \\
\hline
\end{tabularx}
\end{table*}

\section{The GenAI Challenge and Future Research Directions}
The discussed proactive detection methods provide some ways to combat adversarial campaigns, but the evolution of Gen AI still poses challenges.

\subsection{GenAI Attack Swarm}
GenAI enables a new mode of attack: a moving-target-swarm. Instead of a single viral article, adversaries can generate thousands of semantically equivalent but stylistically unique variants of a core narrative \cite{aaai2019}. This inverts the classic assumption of anomaly detection. The standard assumption is that malicious content is a rare class \cite{ieee2024}. However, in a targeted GenAI attack, the swarm of malicious content could form a high-density, low-variance cluster in semantic space \cite{emnlp2025b, ieee2023b}, while the sparse, disconnected pieces of organic content on the topic might appear to be the statistical outliers.

\subsection{Effect of Drifts}
\subsubsection{Multi-level drift}
As discussed earlier, Gen AI enables us to create diverse content at an unprecedented scale. This gives rise to what is known as multi-level drift. Li et al. \cite{arxiv2025b} identify two key drift phenomena:
\begin{itemize}
    \item \textbf{Model-Level Detection Drift:} Even with unchanged underlying semantics, variations in lexical, syntactic or visual aspects may lower model's performance, which is worsened by hallucinations if the detection models are LLM based \cite{arxiv2025b}.
    \item \textbf{Evidence-Level Drift:} Due to such diverse content generated by LLMs, detecting exact matches becomes difficult leading to misleading evidence (especially for diverse news content) \cite{arxiv2025b}.
\end{itemize}
As found out by Li et al. \cite{arxiv2025b}, controlled news diversity (creating diversity through semantically consistent variants of same content by rephrasing or manipulating visual framing) results in misperception drift. Open ended diversity (where Gen AI creates new content by reconstructing narratives or generating images through texts) leads to misclassification.

\subsubsection{Concept Drift}
Shahzad, M. et al \cite{shahzad2025} also define what is known as a concept drift - for any field, domain concepts are susceptible to concept drift where the key attributes change over time, which may force existing detection methods to adapt accordingly. An example Shahzad, M. et al \cite{shahzad2025} listed is that during the corona virus pandemic, facial recognition algorithms had to adapt to recognize masked faces, which was a new attribute for the facial recognition domain. Mathematically, concept drift is defined as
\[ \exists X: P_t (X, y) \neq P_{t+1} (X, y), \]
where $Pt(X, y)$ and $Pt+1(X, y)$ represent the data distribution at time $t$ and $t+1$, respectively \cite{hajmohammed2025}. Concept drift affects the accuracy of models and classifiers, and so is also applicable to LLMs. Deep learning models, including LLMs suffer from concept drift \cite{hajmohammed2025} which can make proactive threat detection more challenging.

\subsubsection{Semantic drift}
Spataru et al \cite{arxiv2024c} also describe semantic drift which points to the decrease in text generation quality when the generation length is increased and is classified as a subtype of hallucinations (direct effect of content instability due to hallucinations cause mentioned earlier in this paper). This happens when the generated text by LLMs diverges from the subject matter specified in the prompts leading to loss of fidelity.

\section{Ethical, Legal and Social Implications}
Analyzing coordination using behavioral and precursor signals can cause ethical challenges. Gupta et al. \cite{mannocci2024} mention that ``The process of collecting and analyzing online data involves monitoring user interactions and behaviors, which can raise concerns about surveillance and consent.''. They also mention that certain coordinated actions might be acceptable in one platform and not in another leading to imbalanced interventions, and that developing methods to silence such coordinated behavior too presents risks as they can be used against targeted or minority groups. This can threaten freedom of speech and expression. Pakina et al \cite{pakina2025} devised a GNN based method to detect synthetic propaganda campaigns. They too have mentioned that their framework that employs behavior analyses of user is subject to implementation challenges which can be due to unintended bias affecting minority or sensitive group of people, misidentifying legitimate users flagged as anomalies, or due to complexities regarding the principle of purpose limitation under European GDPR act (2022).

\section{Proposed Future Research}
\subsection{From Point Anomalies to Cluster Anomalies}
There have been many cluster-based anomaly methods as mentioned in \cite{li2021}, \cite{karami2015} and \cite{last2003}. To detect inauthentic clusters, we need to develop such cluster based techniques in the content space, rather than focusing on point to point behavior. The idea is to aid in detecting patterns emerging from spread of synthetic campaigns based on certain behavior. This should ideally help in proactive detection.

\subsection{Resilience to model level and evidential drift}
Li et al \cite{arxiv2025b} have created a tool known as DRIFTBENCH that examines effect of the mentioned drifts on LVLM based synthetic content detection methods. Their findings emphasize the need for robust, diversity-aware strategies to ensure reliable threat detection.

\subsection{Deep Integration of Behavioral and Content Signals}
In this paper, we explore how proactive detection methods can help within the inauthentic narrative lifecycle. An idea for improving the performance of such models might be to develop signals that integrate proactive behavioral techniques with content based techniques. There has been some prior research on this in the pre LLM era where Vedova et al \cite{vedova2018} have proposed a method that combines news content and social context features that outperforms the methods from that time by increasing detection accuracy by 4.8\%. For video streaming/sharing platforms, this also includes recommendation detection coordination. According to Bu et al \cite{bu2023}, To recommend videos that are of interest to users and are not propagating non-factual artifacts, we need signals to look at the rich side information from recommendation systems and also improve the credibility of recommendations. This can be done by: a) looking at user history to identify topics that the user is more susceptible to and then prioritize those kinds of recommended videos (user-interest-aware detection), b) using user feedback obtained from recommender system (user-feedback-aware detection), and c) add a feature to detect information credibility in recommendation systems (credibility-aware recommendation).

\section{Conclusion}
This survey has systematically reviewed the limitations of the predominantly reactive posture and highlighted the growing body of work enabling a proactive stance of anticipation. The accelerating capabilities of Generative AI have rendered traditional methods insufficient and underscored the critical need for this evolution, primarily by creating a temporal mismatch where falsehoods are generated faster than humans can check them \cite{pnas2024, fhss2025}.

The primary contribution of this work is a systematic review of proactive detection techniques, organized by a unified, lifecycle-based taxonomy, strengthening it by grounding it in a more rigorous, peer-reviewed bibliographic foundation. This was achieved by categorizing current methodologies using the C5 Interaction Model \cite{springer2025} as uniquely suited for proactive monitoring, based on peer-reviewed literature.

The challenges ahead remain formidable. The moving target swarm and multi-level drift problems demand a corresponding step-change in our defensive strategies. However, the future research agenda is now more defined. The synthesis of these components, such as spotting the Coordinated Seeding, identifying Anomalous Propagation, and feeding those signals into an AaaS-driven system, provides a cohesive roadmap.

The research community can move beyond an endless cycle of cleaning up synthetic noise by adopting this integrated, interdisciplinary, and anticipatory paradigm, which unites technical innovation with profound ethical humility. The goal is not simply to engineer information ecosystems that are resilient by design, but to ensure they are also responsible, transparent, and rights-preserving.

\begin{IEEEbiographynophoto}{JONGHYUN CHUNG} is a Software Engineer at Google with over five years of experience in machine learning and research at different companies, including Google and Samsung. He holds Master's and Bachelor's degrees in Computer Science from Cornell University, where he specialized in Natural Language Processing. At Google, his work focuses on backend systems and building production-grade large-scale machine learning systems.
\end{IEEEbiographynophoto}

\begin{IEEEbiographynophoto}{RISHABH CHADDHA} is a software engineer at Google with experience in full stack and ml infra development across YouTube Shopping, YouTube Responsibility, and YouTube News teams. He has a Masters in Computer Science from Purdue, Bachelors in Computer Science and Engineering from BIT Mesra (India) with a 5 year industry experience across different parts of the software stack.
\end{IEEEbiographynophoto}

\begin{IEEEbiographynophoto}{SANKET BADHE} is a seasoned Machine Learning Engineer with over 10 years of experience specializing in AI Security, large-scale ML systems, and LLM applications. He currently leads key ML initiatives at Google for Youtube shopping. Sanket holds a Master’s in Data Science from Rutgers University and a B.Tech from IIT Roorkee, with prior experience at Tinder, TikTok, Oracle etc. His research has been published in ACL, CAMLIS and IEEE.
\end{IEEEbiographynophoto}

\begin{IEEEbiographynophoto}{DEBANSHU DAS} is a Senior Software Engineer and Technical Lead at Google, where he focuses on generative-AI driven solutions for large-scale recommendation systems in the context of creative advertising on YouTube. His work spans generative AI, agentic workflows, large-scale recommendation engines, and cloud-native distributed systems. He brings prior industry experience from Oracle and Apple, enabling him to bridge theoretical research and production-grade system design. An IEEE Senior Member, his research has been accepted at leading venues including AAAI and WSDM. He holds a Master’s degree from Carnegie Mellon University.
\end{IEEEbiographynophoto}

\begin{IEEEbiographynophoto}{NATHAN HUANG} is a Data Scientist at Google with over a decade of experience spanning statistical sampling, machine learning, econometrics and quantitative marketing. He holds a Ph.D. in Quantitative Marketing and an M.S. in Economics from the University of Wisconsin-Madison, as well as an undergraduate degree in Economics from the University of International Relations.
\end{IEEEbiographynophoto}

\begin{IEEEbiographynophoto}{AMANPREET KAUR} is a Senior Engineering Analyst at Google, bringing over 11 years of data science leadership, spearheading advancements within Machine Learning and Generative AI integration for key Google Search products. With a foundational BTech in Computer Science and a history in data scientist and analytics roles, her expertise includes quantitative metric design and instrumentation, rigorous experiment design, advanced statistical modeling \& feature engineering leveraging large-scale logs, and the development, tuning, and evaluation of ML and LLM based classifiers and algorithms.
\end{IEEEbiographynophoto}
\EOD

\end{document}